%% file: rui.tex
\documentclass[lettersize,journal]{IEEEtran}
\usepackage{amsmath,amsfonts}
\usepackage{algorithmic}
\usepackage{algorithm}
\usepackage{array}
\usepackage[caption=false,font=normalsize,labelfont=sf,textfont=sf]{subfig}
\usepackage{textcomp}
\usepackage{stfloats}
\usepackage{url}
\usepackage{verbatim}
\usepackage{graphicx}
\usepackage{cite}
\usepackage{enumitem}
\usepackage[table,xcdraw,dvipsnames]{xcolor}

\usepackage{hyperref}
\usepackage{etoolbox}
\makeatletter
\patchcmd{\@makecaption}
  {\scshape}
  {}
  {}
  {}
\patchcmd{\@makecaption}
  {\\}
  {.\ }
  {}
  {}

\begin{document}

\title{Continual Deep Active Learning for Medical Imaging: Replay-Base Architecture for Context Adaptation}

\author{Rui Daniel, M. Rita Verdelho, Catarina Barata, Carlos Santiago

Instituto de Sistemas e Robótica, Instituto Superior Técnico - Universidade de Lisboa, Portugal

Email: ruipcdaniel@tecnico.ulisboa.pt
}



\maketitle

\begin{abstract}
\input{sections/S0_abstract}

\end{abstract}


\input{sections/S1_introduction}
\input{sections/S2_background}

\input{sections/S3_methodology}

\input{sections/S4_experimental_setup}
\input{sections/S5_experimental_results}

\input{sections/S6_conclusions}
\bibliographystyle{IEEEbib}
\bibliography{strings,refs}

\appendix

\input{sections/attachments}

\vfill

\end{document}

%% file: sections/S0_abstract.tex
Deep Learning for medical imaging faces challenges in adapting and generalizing to new contexts. Additionally, it often lacks sufficient labeled data for specific tasks requiring significant annotation effort. 
Continual Learning (CL) tackles adaptability and generalizability by enabling lifelong learning from a data stream while mitigating forgetting of previously learned knowledge. 
Active Learning (AL) reduces the number of required annotations for effective training.
This work explores both approaches (CAL) to develop a novel framework for robust medical image analysis.
Based on the automatic recognition of shifts in image characteristics, Replay-Base Architecture for Context Adaptation (RBACA) employs a CL rehearsal method to continually learn from diverse contexts, and an AL component to select the most informative instances for annotation. 
A novel approach to evaluate CAL methods is established using a defined metric denominated IL-Score, which allows for the simultaneous assessment of transfer learning, forgetting, and final model performance.
We show that RBACA works in domain and class-incremental learning scenarios, by assessing its IL-Score on the segmentation and diagnosis of cardiac images. 
The results show that RBACA outperforms a baseline framework without CAL, and a state-of-the-art CAL method across various memory sizes and annotation budgets.
Our code is available in \href{https://github.com/RuiDaniel/RBACA}{https://github.com/RuiDaniel/RBACA}.

%% file: sections/S1_introduction.tex
\section{Introduction}

\label{chap:intro}

The complexity of healthcare interventions, from diagnosis to treatment planning through segmentation, presents significant challenges, creating the need for computational systems assistance. Deep Learning (DL) has been explored for classification, detection, and segmentation tasks, and its learning capacity is already impacting computer-assisted healthcare applications. However, DL in the medical field is still unreliable. Current models are unable to adapt and generalize to new contexts, such as different scanner settings, and there is a lack of labeled data for specific tasks, which is hard to obtain.


To tackle adaptability and generalizability to new contexts, Transfer Learning \cite{pan2009survey} develops models capable of generalizing from a source domain and task to a target domain and task. In contrast, Continual Learning (CL) \cite{van2022three} involves incremental learning from a data stream across diverse contexts without a predefined target context. It aims to mitigate catastrophic forgetting, which occurs when models trained on new data tend to lose previously acquired capabilities. Therefore, CL is more suitable for adaptability to multiple and varying contexts.

To minimize the cost of obtaining labeled data, Active Learning (AL) \cite{settles1994active} methods can be considered when unlabeled data is abundant, as in the case of continuous data streams. AL aims to minimize the amount of data required for effective training by identifying the most informative samples for labeling. The labeling process is performed by an oracle in annotation rounds (query rounds).
In pure AL, models are retrained from scratch at each query round, which has high time complexity.
Fine-tuning in each round may lead to catastrophic forgetting, while incorporating CL enables the incremental learning of new information. The fusion of the two methods is called Continual Active Learning (CAL) \cite{vu2023active}.


The objective of this work is to develop tools that enhance the capacity of DL algorithms to adapt and generalize to diverse contexts. Several contributions were achieved:

\begin{itemize}
    \item A novel CAL framework, denominated Replay-Based Architecture for Context Adaptation (RBACA), was developed for medical image analysis. It employs a CL memory-based rehearsal method and supports various memory management and memory pruning strategies.
    
    \item RBACA promotes efficient selection of samples for annotation by considering both informativeness and diversity within an annotation budget.
    
    \item RBACA is adaptable to incremental learning of domains and classes, as demonstrated through segmentation and diagnosis of cardiac images.
    
    \item A novel approach to evaluate CAL methods was established using a defined metric, termed Incremental Learning Score (IL-Score), which simultaneously assesses transfer learning, forgetting, and final model performance.
   
    \item RBACA outperforms both a baseline framework without CAL and a state-of-the-art CAL method across various memory sizes and annotation budgets. RBACA enhances adaptation and generalization to diverse medical imaging contexts by improving the quality of memory representations.
    
\end{itemize}

The remainder of the paper is organized as follows: Section 2 reviews the background, Section 3 introduces the methodologies used, Section 4 describes the experimental setup, Section 5 presents the results, and Section 6 concludes the paper.

%% file: sections/S2_background.tex
\section{Related Work}

\label{stateofART}

The performance of many DL techniques relies on both training and test sets being drawn from the same distribution, which is unlikely in the real world.
When such a shift in distribution occurs, a strategy is to rebuild models from scratch using newly collected training data, which can be unfeasible.

CL \cite{van2022three} involves adapting models to new contexts while maintaining performance on previously learned ones, without retraining from scratch at each new learning phase.
There are three fundamental CL scenarios \cite{van2022three}: Domain-Incremental Learning (Domain-IL), Task-Incremental Learning (Task-IL) and Class-Incremental Learning (Class-IL). 

Task-IL involves incrementally learning a set of clearly distinct contexts referred to as tasks. At all times task identity must be known. This approach allows training models with task-specific components or with distinct networks for each task. The main challenge lies in effectively sharing learned representations across tasks.
Domain-IL involves incrementally learning a particular task across various contexts designated domains. At test time, the algorithm does not require domain identification, as all of them share the same potential outputs.
Class-IL involves learning to discriminate among an increasing number of classes, which appear as contexts denominated episodes. Consequently, a challenge is learning to discriminate between classes that have not been observed together, as different episodes can contain different classes.

According to De Lange \textit{et al.} \cite{de2021continual}, CL strategies can be categorized into 3 families: memory-based, regularization-based, and parameter isolation methods. However, certain methods utilize techniques from multiple families.
Memory and replay mechanisms use storage to recall past contexts, combining new data with previous representations when learning a new context. They face challenges with memory requirements and privacy concerns, as data access is often restricted in real-world situations, such as in health applications where legislation limits long-term data storage \cite{masana2022class}.

These techniques fall into three categories: rehearsal, pseudo-rehearsal, and constrained optimization. 
Rehearsal methods involve replayed stored samples. However, they may lead to overfitting on the replayed subset. In van de Ven \textit{et al.} \cite{van2022three}, Experience Replay (ER) involves replayed data chosen uniformly at random from previous contexts. Zheng \textit{et al.} \cite{zheng2024selective} adapted ER to address brain tumor segmentation. 

The memory may also contain latent representations. For instance, Incremental Classifier and Representation Learning (iCaRL) \cite{rebuffi2017icarl} learns strong classifiers and representations by storing templates closest to the feature mean of each class. 
Pseudo-rehearsal methods involve generating pseudo-samples using a generative model.
For instance, Deep Generative Replay (DGR) \cite{shin2017continual} replays generated representations at the input level. 
Constrained optimization allows for greater freedom in transferring knowledge between tasks. Under Task-IL, Averaged Gradient Episodic Memory (A-GEM) \cite{chaudhry2019efficient} enforces the inequality constraint that the loss on replayed data cannot increase during optimization of the loss on new data.

Regularization-based methods involve adding a regularization term to the loss. This term penalizes changes to important parameters that represent previously learned contexts.
Data-focused techniques rely on functional regularization, meaning that when training on new contexts, they aim to retain the previously learned input-output mapping at specific anchor points. For instance, Learning without Forgetting (LwF) \cite{li2017learning} uses current context inputs as anchor points and has been applied to classifying retinal diseases in Verma \textit{et al.} \cite{verma2023privacy}.
In contrast, prior-focused methods estimate parameter importance distribution. For example, Elastic Weight Consolidation (EWC) \cite{kirkpatrick2017overcoming} quantifies parameter importance by utilizing the Fisher information matrix and has been applied to classifying optical coherence tomography in Verma \textit{et al.} \cite{verma2023privacy}.

Finally, parameter isolation methods employ distinct model parameters for each context.
In Fixed Networks, the architecture remains static, with fixed context-specific components to each context. 
In contrast, Dynamic Architectures do not impose constraints on architecture size. They involve adding new neurons to models as they encounter new contexts, which allows for knowledge preservation by updating both old and newly added parameters at a different learning pace \cite{wang2017growing}. 

AL strives for high accuracy with a minimal number of annotations by employing methods to select the most informative samples for labeling. The selection process involves a trained model from the labeled pool $L$. Newly labeled instances are added to $L$, and the model learns from them. In contrast to serial selection, batch-mode selection allows querying instances in groups, which enhances labeling efficiency.

In the research conducted by Settles \textit{et al.} \cite{settles1994active}, three AL scenarios are introduced: Membership Query Synthesis, Stream-Based Selective Sampling, and Pool-Based Active Learning.
In the first one, the learner may request labels for any unlabeled instance within the input space, including queries it generates, which can lead to outliers.
On the other hand, in Stream-Based Selective Sampling, the learner decides whether to label or discard each sequentially sampled unlabeled instance. 
Finally, in Pool-Based Active Learning, a pool of unlabeled data is evaluated to select the best query.

Settles \textit{et al.} \cite{settles1994active} presented several methods for selecting the most informative samples to be labeled. 
Uncertainty Sampling \cite{lewis1995sequential} selects instances where the model is least confident about the labeling, with entropy \cite{shannon1948mathematical} being the commonly used uncertainty metric. 
Query-By-Committee (QBC) \cite{seung1992query} employs a committee of models, all trained on the current labeled set. The models vote on the labeling of each unlabeled instance, identifying the samples where they disagree the most. Grimova \textit{et al.} \cite{grimova2019query} applied both QBC and Uncertainty Sampling to the classification of overnight polysomnography records.

In contrast, Expected Gradient Length (EGL) \cite{settles2007multiple} selects the instance whose labeling would result in the greatest training gradient. However, as the true label is unknown, the gradient is computed as an expectation across potential labelings, which is computationally demanding.
Another method is Estimated Error Reduction, which directly aims to minimize generalization error. However, for each sample, the model must be incrementally re-trained for every potential labeling.

Batch-mode selection aims to assemble the optimal query batch. Some strategies focus on informativeness, such as Bayesian Active Learning by Disagreement (BALD) \cite{houlsby2011bayesian} which was applied by Gal \textit{et al.} \cite{gal2017deep} for skin cancer diagnosis. However, this may select a set of informative yet similar samples. To more accurately represent the input distribution, other strategies integrate diversity-based approaches. On the other hand, hybrid query strategies consider both informativeness and diversity. For example, BatchBALD \cite{kirsch2019batchbald} extends BALD by incorporating the correlation between batch data points.

DL can automatically learn and extract relevant features from imaging annotated data, leveraging techniques such as Convolutional Neural Networks (CNNs). Deep Active Learning \cite{ren2021survey} aims to minimize annotation costs while maintaining DL capabilities.
A DL model typically consists of two stages: feature extraction and task learning. Consequently, the uncertainty of the DL model encompasses the uncertainty inherent in both stages. For instance, Learning Loss for Active Learning (LLAL) \cite{yoo2019learning} leverages the output of multiple hidden layers as input for a loss prediction module, which guides the query strategy.
DL requires a substantial volume of labeled data for effective training. One strategy for expanding the labeled sample dataset involves assigning pseudo-labels to high-confidence samples, as in Cost-Effective Active Learning (CEAL) \cite{wang2016cost}.

AL presents several challenges, such as a limited labeling budget and a lack of adaptability. Models are typically retrained from scratch in each query round, using the ever-increasing labeled pool, which is computationally inefficient. Re-training and fine-tuning at each round is not a viable solution due to catastrophic forgetting.
CL facilitates adaptation to future domains while minimizing the risk of forgetting. However, it requires all data to be labeled. These AL and CL disadvantages give rise to CAL, wherein the model progressively learns new information in each round of annotations. 


CAL methods have demonstrated the capability to surpass AL techniques, achieving noteworthy speedups. 
Das \textit{et al.} \cite{das2023accelerating} introduces Scaled Distillation w/ Submodular Sampling (CAL-SDS2). This method regularizes the model and utilizes submodular functions to capture notions of diversity and representativeness while achieving a considerable speedup in a dermatoscopy dataset.
Ayub \textit{et al.} \cite{ayub2022few} present Few-Shot Continual Active Learning (FoCAL) for object classification, with a CL pseudo-rehearsal strategy and an AL uncertainty sampling.
Vu. \textit{et al.} \cite{vu2023active} explore CAL in Domain-IL, Task-IL and Class-IL scenarios. They propose a forgetting-learning profile to understand the relationship between the contrasting goals of low forgetting and quick learning ability.

Continual Active Learning for Scanner Adaptation (CASA), a CAL architecture developed by Perkonigg \textit{et al.} \cite{perkonigg2022continual}, is designed to process a continuous stream of images in a multi-scanner environment. It operates within a Domain-IL CL scenario using a replay-based rehearsal method where annotations are stored in the rehearsal memory. An AL algorithm automatically detects shifts in image characteristics, selects examples for labeling within a limited budget, and adapts training accordingly.

Several concerns have been identified with replay-based architectures. The quality of memory representations is affected by factors such as memory size, annotation budget, memory management strategies, memory pruning techniques, and AL methods.
For instance, in Vu. \textit{et al.} \cite{vu2023active}, a random sampling technique is used both as an AL method and as a pruning strategy.
Furthermore, FoCAL focuses on achieving a good trade-off between memory storage and model performance.
On the other hand, CASA uses Least Recently Used (LRU) \cite{belady1969anomaly} pruning which may remove crucial data. Moreover, its AL method relies on a performance metric (Perf) on similar stored samples, which may disregard highly informative samples.

%% file: sections/S3_methodology.tex
\section{Methodology}
\label{chap:proposta}

The identified concerns with the state-of-the-art, particularly replay-based methods, have motivated the development of a novel framework called Replay-Based Architecture for Context Adaptation (RBACA). This framework aims to improve the quality of context representations in memory, thereby enhancing performance.
RBACA employs a CL memory-based rehearsal method and an AL component that selects informative instances. 

\begin{figure}[t]
\begin{minipage}[b]{\linewidth}
  \centering
  \centerline{\includegraphics[width=1\linewidth]{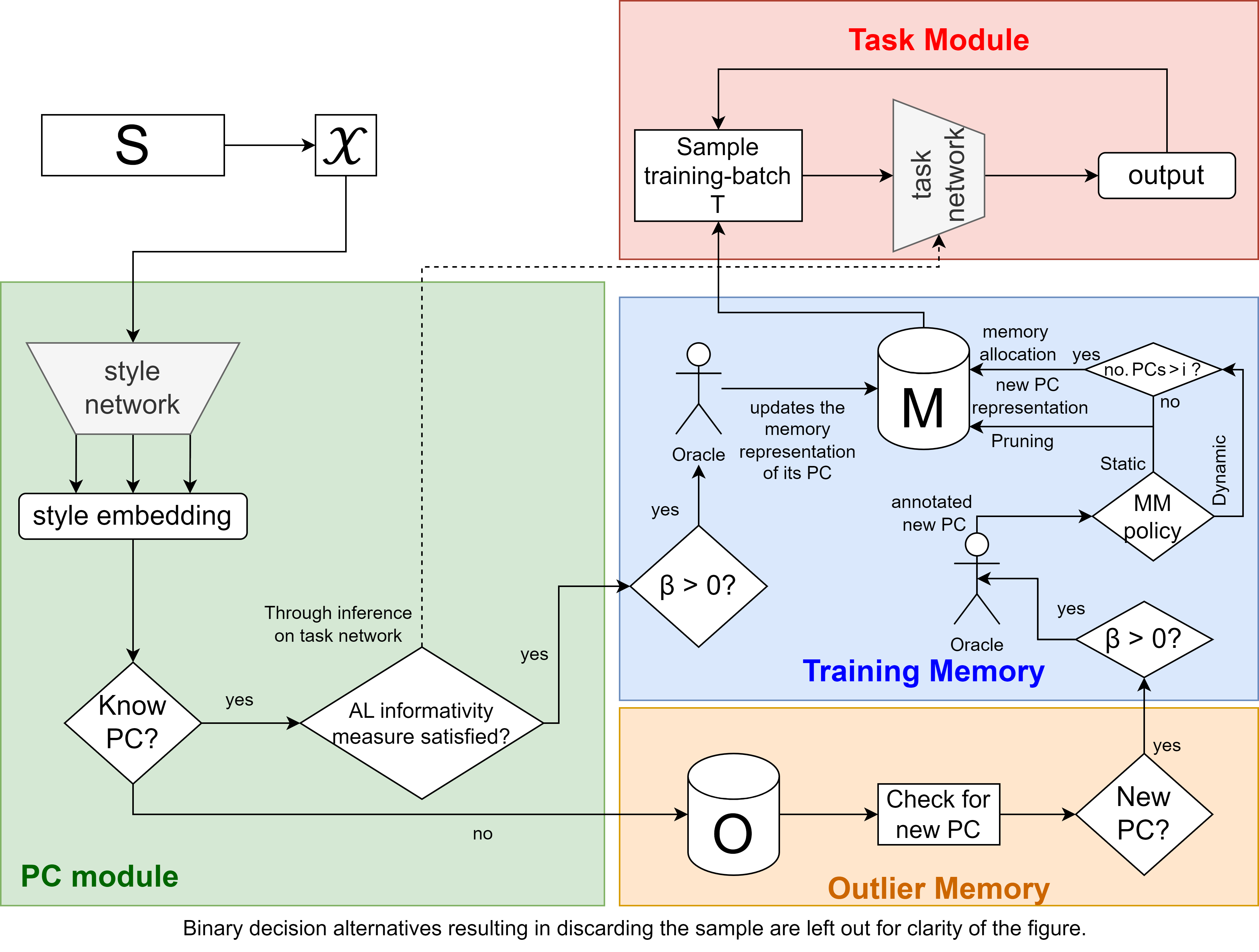}}
\end{minipage}
\caption{The RBACA architecture. Processes a data stream $S$ and is composed of four key components: PC module, outlier memory $O$, training memory $M$, and task module. It supports various memory management (MM), pruning, and AL strategies.}
\label{fig:mymodel}
\end{figure}

Fig. \ref{fig:mymodel} shows the RBACA architecture. The Pseudo-Context (PC) module evaluates each data stream sample, deciding whether to route it to the oracle or to the outlier memory $O$ (AL method). When $O$ receives a new item, an evaluation determines whether a new PC should be created and then directed to the oracle. The oracle labels and stores these instances in the rehearsal training memory $M$ of size $K_M$, which is utilized by the task module for network training (CL component).

We build over the CASA \cite{perkonigg2022continual} architecture, improving memory management, memory pruning, identification of relevant samples, and expanding the model to Class-IL scenarios.
Both task network and style network can be designed based on the desired task (\textit{e.g.}, segmentation or diagnosis).
Compared to CASA, RBACA employs more complex techniques; however, this is masked by its parallelizability.


\subsection{Memory Management}

$M$ is balanced among existing Pseudo-Contexts (PCs). Upon the addition of a new PC, pruning is executed for each PC to maintain balance. Both labeling budget $\beta$ and $K_M$ influence the diversity of PCs in $M$ and the quality of their representations. 
Early consumption of $\beta$ can result in the non-representation of later PCs in memory. With a small $K_M$, memory saturation may occur, compromising representations, especially as the number of PCs increases. On the other hand, an excessively large $K_M$ may lead to an imbalance in $M$ between PCs.
These concerns can be addressed by enabling memory management (MM) to shift from Static to Dynamic. 

When a new PC is found, its samples are annotated and added to $M$. $M$ must have space for this new set. In Static MM, when a new PC is found, pruning is applied to all existing PCs. In contrast, in Dynamic MM, $k$ new elements are allocated for each newly detected and annotated PC until the maximum allowed system memory is reached. However, an initial memory size of only $k$ may be insufficient for effective training. Therefore, in a dynamic DM-$i$ strategy, pruning occurs for the first $i-1$ newly detected PCs.

\subsection{Memory Pruning}

Dataset pruning, without synthesizing representations, summarizes the dataset into a subset valuable for training by selecting which samples to retain and which to remove. Multiple pruning techniques were considered, as their performance varies with data type and volume. One of these is LRU pruning, as employed by Perkonigg \textit{et al.} \cite{perkonigg2022continual}.

Among distribution-based techniques, K-Means \cite{macqueen1967some}, Gaussian Mixture Models (GMMs) \cite{reynolds2009gaussian}, and Density-Based Spatial Clustering of Applications with Noise (DBSCAN) \cite{ester1996density} strategies are compared. In these approaches, the samples selected for retention are those closest to each cluster's centroid.
K-Means groups samples by assigning each point to the nearest cluster centroid, resulting in low computational complexity. However, it is sensitive to noise, does not account for variance, and requires the number of clusters in advance. In contrast, GMMs can handle clusters of arbitrary shapes by computing the probabilities of each point belonging to each cluster. Finally, DBSCAN provides a density-based approach to clustering that does not require specifying the number of clusters.

Informativeness-based techniques: Uncertainty Sampling \cite{lewis1995sequential} and EGL \cite{settles2007multiple} were also considered.
Additionally, several hybrid strategies that balance diversity and informativeness were tested, such as K-Means combined with Uncertainty Sampling (KU) and EGL combined with GMMs (EGLGMM). In these approaches, clustering is first applied. For each cluster $i$, $k_i$ samples are selected: the $k_i/2$ closest to the centroid, and among the remaining samples, the $k_i/2$ most informative.

\subsection{Informativeness-Based Query Method}

When a new sample belongs to a known PC, an AL algorithm must determine whether to annotate or discard the instance. To make this decision, RBACA enables an uncertainty sampling strategy based on informativeness, using a decision threshold. The uncertainty of each sample is calculated by inferring the unlabeled data pool through the task network.

\subsection{Adaptation to Class-IL}

RBACA is adaptable for Class-IL. This scenario begins with a classification problem within a single episode. Then, by leveraging shared learned representations across episodes, the number of classes expands.
RBACA for Class-IL involved modifications in the data loading phase, the task network model, and the metrics. The most crucial adaptation occurs in the final layer of the task network. This layer initially starts with 0 units. During each training step, when a new class is detected, a corresponding unit is added to the final layer while preserving the weights of the already identified classes.
Therefore, RBACA is adaptable to any classification problem, as the number of classes is not predefined, and new classes may emerge.

%% file: sections/S4_experimental_setup.tex
\section{Experimental Setup}

\label{chap:es}

This section describes the dataset used,
tasks, training conditions, data manipulation techniques, metrics, and baselines.

\subsection{M\&Ms Dataset}
We evaluate the performance of RBACA in cardiac imaging problems using the Multi-Centre, Multi-Vendor, and Multi-Disease Cardiac Image Segmentation Challenge dataset (M\&Ms) \cite{campello2021multi}.
Our model is applied to Domain-IL (segmentation across multiple scanners and centers) and Class-IL (multiple disease diagnosis across multiple scanners and centers) problems.
M\&Ms includes cardiovascular magnetic resonance segmented 2D images with contours for the background (BG), left ventricle (LV), left ventricular myocardium (MYO), and right ventricle (RV). 
These correspond to segmentation classes 0, 1, 2, and 3, respectively.
Each patient examination is recorded in a separate file, including an external code, vendor, center, End-Diastolic (ED) frame, and End-Systolic (ES) frame. Each file contains multiple frames, each featuring several slices. The ED frame represents the heart when the ventricles are fully filled with blood. The ES frame represents the heart when the ventricles have completed contraction. So both of these frames include the most relevant slices.

M\&Ms includes the following pathologies: Hypertrophic Cardiomyopathy (HCM) - 24.6\%; Dilated Cardiomyopathy (DCM) - 28.1\%; Hypertensive Heart Disease (HHD) - 7.0\%; Abnormal Right Ventricle (ARV) - 4.3\%; Athlete Heart Syndrome (AHS) - 0.9\%; Ischemic Heart Disease (IHD) - 1.4\%; Left Ventricle Non-Compaction (LVNC) - 0.6\%; Healthy/within normal limits (NOR) - 25.8\%; and Other - 7.2\%. 
The dataset comprises 4 vendors across 5 centers: C1 - Hospital Vall d'Hebron (Barcelona) - Siemens; C2 - Clínica Sagrada Familia (Barcelona) - Philips; C3 - Universitätsklinikum Hamburg-Eppendorf (Hamburg) - Philips; C4 - Hospital Universitari Dexeus (Barcelona) - General Electric; and C5 - Clínica Creu Blanca (Barcelona) - Canon. 

Patient-level splitting divided the dataset into a base training set (1130 slices), a continual training set (4322 slices), a validation set (922 slices), and a test set (1544 slices).
The continual training set initially consists of unlabeled data, with the M\&M centers appearing in the order 1, 4, 2, 3, and 5, based on Perkonigg \textit{et al.} \cite{perkonigg2022continual}.
The oracle provides annotations by revealing the labels.




\label{TasksandTrainingConditions}
\subsection{Tasks and Training Conditions}
ResNet-50 \cite{ren2016faster}, a CNN pretrained on ImageNet, was used as style network across all tasks.
The training conditions for both tasks included a batch size of 8, Adam optimizer with a learning rate of $10^{-4}$, 10 epochs for base training, and 3 epochs of training on the rehearsal memory, where a training step is performed whenever the number of updates without training
exceeded 9. To ensure reproducibility in the experiments, a seed must be selected for each run, ensuring deterministic behavior across both CPU and multi-GPU setups.  


Using M\&Ms, a cardiac segmentation task was defined, where the objective is to classify each pixel. During model training, the background class (BG) is excluded from the loss calculation to
ensure that the large number of background pixels does not bias the training process. A 2D-UNet \cite{ronneberger2015u} was used as task network.
This task was addressed within Domain-IL, and various loss were tested: Cross-Entropy (CE), Focal Loss (FL) \cite{lin2020focal}, Dice, DiceCE, and DiceFL.




Cardiac pathology classification task is defined as a multi-class problem under Class-IL, where the model learns new categories as new
samples from distinct centers become available. M\&Ms, with 9 categories, is well-suited for this, as different centers contain distinct pathologies.
ResNet-50 and Vision Transformer (ViT) were tested. 
Based on Dosovitskiy \textit{et al.} \cite{dosovitskiy2020image}, the selected ViT is a transformer encoder pre-trained on ImageNet-21k and fine-tuned on ImageNet 2012, both at a resolution of 224$\times$224.
While ResNet-50 is effective with limited data and computational resources, ViT can potentially outperform it in cases with large datasets and substantial computational power. Both models were tested using CE loss.

In both tasks, the number of clusters for K-Means was set to 5, and the number of components for GMMs was set to 5. Additionally, DBSCAN is considered, with a maximum distance of 0.1 between neighbors and a minimum of 3 neighbors required for a point to be considered a core point.






\label{dm}

\subsection{Data Processing}
For ResNet-50 and UNet, the images were center-cropped to 240$\times$196 pixels and normalized to a [0,1] range. 
In contrast, the used ViT has specific preprocessing requirements. Images were resized to 224$\times$224 pixels and normalized across the RGB channels with a mean and standard deviation of 0.5 per channel. \footnote{We used the \texttt{vit-base-patch16-224} by Hugging Face, with the preprocessing procedure described \href{https://huggingface.co/google/vit-base-patch16-224}{here}.}
To enhance generalization and performance, online data augmentation (DAug) was employed during training. In each instance, RBACA randomly applies one of ten possible transformations: no augmentation; 180-degree rotation; horizontal and vertical flips; Gaussian blur with a kernel size of (0.06, 0.06) and a standard deviation of 0.006; brightness increase/decrease by 10\%; contrast increase/decrease by 10\%; and Gaussian noise addition with a mean of 0 and a standard deviation of 0.035. 



\label{metrics}

\subsection{Evaluation}
For segmentation, the Dice Similarity Coefficient (DSC) assesses agreement between prediction and target. In classification, F1 Score is calculated as the arithmetic mean of the per-class metrics.
To assess forgetting, Backward Transfer (BWT) \cite{lopez2017gradient} measures how learning a new context affects performance on previously learned contexts. BWT is defined as 
\begin{equation}
\text{
    $\text{BWT} = \frac{1}{t-1} \sum_{j=1}^{t-1} (a_{t,j} - a_{j,j})$ \; ,
}
\label{bwt}
\end{equation}
where $a_{x,y}$ represents the test metric (DSC or F1 Score) for context $y$ after learning from context $x$, and $t$ is the final learned context. Conversely, Forward Transfer (FWT), which is given by
\begin{equation}
\text{
    $\text{FWT} = \frac{1}{t-1} \sum_{j=2}^{t} (a_{j-1,j} - \overline{b_j})$ \; ,
}
\label{fwt}
\end{equation}
quantifies the impact of prior contexts on the performance of new contexts. Here, $\overline{b_j}$ denotes the test metric for context $j$ at random initialization. The relevant metrics are DSC/F1 Score (range [0,1]), BWT (range [$-$1,1]), and FWT (range [$-$1,1]). Therefore, we propose a novel approach to evaluate CAL methods by using the average of these three metrics, denominated as Incremental Learning Score (IL-Score). This metric allows for the simultaneous evaluation of transfer learning, forgetting, and final model performance. IL-Score ranges from [$-$2/3, 1], where higher values indicate superior performance.




\label{baselines}

\subsection{Model Comparison}

Various baselines were established. 
\textbf{ContextEval} aims to show the impact of prior contexts on new contexts' performance. Assuming there are $C$ contexts, it involves using $C-1$ contexts for training and one for testing in each round. This process is repeated $C$ times, rotating the test context. Finally, the average FWT across rounds is calculated.
\textbf{SeqFineTune} enables to assess forgetting and transfer learning, in a framework without CAL. The model is fine-tuned with each context sequentially, resulting in $C$ fine-tuning procedures. FWT, BWT and task-specific metrics are obtained. In this study, each center represents a context, so $C$ is equal to the number of centers in M\&Ms, $C$=5.
The performance of RBACA was also compared with a state-of-the-art CAL method (CASA).

%% file: sections/S5_experimental_results.tex
\section{Experimental Results and Discussion}

\label{chap:results}

This section presents and discusses the experimental results through both quantitative and
qualitative analyses. For each experiment, seeds 1, 2, and 3 were used, and the results are presented as the mean ± standard deviation.

\subsection{Cardiac Segmentation}
\label{CardiacSegmentationResultsandDiscussion}

For the cardiac segmentation task, results were obtained without DAug as in Perkonigg \textit{et al.} \cite{perkonigg2022continual}, to enable a direct comparison with CASA.
Various loss functions were evaluated using RBACA, with $\beta=430$, $K_M=128$, Static MM, Perf AL method and LRU pruning. DiceCE achieved the highest IL-Score and was selected.

To determine the best uncertainty threshold, $U_{th}$, for the AL method, a study was conducted using the same parameters. The results are reported in Table \ref{al-tuning}.
$U_{th}=0.025$ was identified as the best configuration and serves as a reference for subsequent tests. At this threshold, the annotation budget is used ideally because it depletes as late as possible. This allows for new annotations to be made continuously until the end of the stream, leading to a more representative memory. As a result, retraining occurs more frequently. 

\begin{table}[t]
\centering
\caption{AL $U_{th}$ with $\beta=430$, $K_M=128$, Static MM, and LRU pruning. Label counter is the number of annotations, and train counter is the number of batch training steps performed.}
\label{al-tuning}
\begin{tabular}{|c|c|c|c|}
\hline
\rowcolor[HTML]{B7B7B7} 
\begin{tabular}[c]{@{}c@{}}AL \\ Method\end{tabular} & \begin{tabular}[c]{@{}c@{}}Label \\ Counter\end{tabular} & \begin{tabular}[c]{@{}c@{}}Train \\ Counter\end{tabular} & IL-Score                    \\ \hline
\rowcolor[HTML]{EFEFEF} 
$U_{th}$=0.001                                            & 430                                                      & 140                                                      & 0.426 ± 0.003          \\ \hline
$U_{th}$=0.01                                             & 430                                                      & 269                                                      & 0.418 ± 0.008          \\ \hline
\rowcolor[HTML]{EFEFEF} 
$U_{th}$=0.02                                             & 430                                                      & 1136                                                     & 0.434 ± 0.006          \\ \hline
$U_{th}$=0.0225                                           & 273                                                      & 879                                                      & 0.426 ± 0.004          \\ \hline
\rowcolor[HTML]{EFEFEF} 
$U_{th}$=0.025                                            & 427                                                      & 1786                                                     & \textbf{0.437 ± 0.002} \\ \hline
$U_{th}$=0.0275                                           & 234                                                      & 877                                                      & 0.423 ± 0.011          \\ \hline
\rowcolor[HTML]{EFEFEF} 
$U_{th}$=0.03                                             & 64                                                       & 860                                                      & 0.424 ± 0.007          \\ \hline
$U_{th}$=0.05                                             & 52                                                       & 67                                                       & 0.427 ± 0.002          \\ \hline
\end{tabular}
\end{table}

Using ContextEval, the results are DSC = 0.563 ± 0.003 and FWT = 0.538 ± 0.016. 
FWT is calculated as the difference between DSC and DSC obtained from a randomly initialized untrained model. The significant FWT observed demonstrates that training on a new context benefits when the model has been previously trained on earlier contexts. 

SeqFineTune overall results are reported in Table \ref{fine-tune-seg-comparison}.
The significant FWT indicates a substantial benefit in using transfer learning. However, the negative BWT shows evidence of forgetting, meaning that, generally, the performance for a center decreases after learning from other contexts.

Given that real-world problems often involve multiple constraints, such as limited annotation budgets or restricted system memory, nine configurations were considered. These configurations combine the labeling budget $\beta = \{108, 430, 860\}$ with the memory size $K_M = \{200, 485, 2000\}$, representing low, medium, and high values for each variable. There are 4322 training slices in the M\&Ms dataset. Thus, the $\beta$ values approximately correspond to $\{2.5\%, 10\%, 20\%\}$ of the training set size, while $K_M$ represents approximately $\{4.6\%, 11.2\%$, and $46.3\%\}$ of the training set size.
These values were selected within the training set size to evaluate the performance of the methods under various configurations of $\beta$ and $K_M$.



For each combination of $\beta$ and memory size, RBACA was tested with MM: \{Static, DM-1, DM-2, DM-3\}; Pruning: \{LRU, K-Means, GMMs, DBSCAN, Uncertainty, EGL, KU, EGLGMM\}; and AL: \{Perf, $U_{th} = \{0.025, 0.0225, 0.02\}\}$.

With Static MM, LRU pruning, and Perf AL method, on average, around 5 PCs are formed.
Consequently, $K_M = \{200, 485, 2000\}$ will, on average, correspond to a final
memory size per PC of $k = \{40, 97, 400\}$. In Static MM, $K_M$ is fixed and $k$ depends on the number of PCs. In contrast, in Dynamic MM, $k$ is fixed and $K_M$ depends on the number of PCs.
Therefore, when applying Dynamic MM, $k$ will be used as the defined parameter instead of $K_M$.

\begin{figure}[t]
\begin{minipage}[b]{\linewidth}
  \centering
  \centerline{\includegraphics[width=1\linewidth]{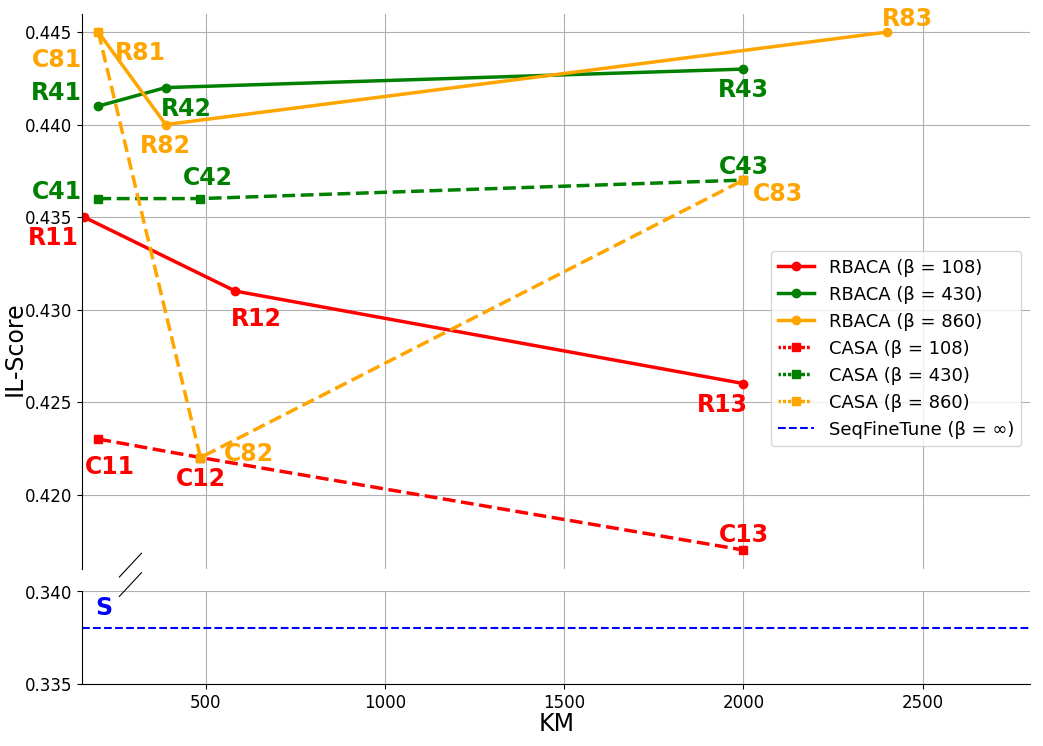}}
\end{minipage}
\caption{RBACA, CASA and SeqFineTune segmentation IL-Score for multiple configurations. Points are labeled using the XYZ format, where X denotes either R or C, representing RBACA or CASA; Y indicates 1, 4, or 8, corresponding to $\beta$ values of 108, 430, or 860; and Z represents 1, 2, or 3, relating to $k$ values of 40, 97, or 400 in Dynamic MM, or to $K_M$ values of 200, 485, or 2000 in Static MM.}
\label{fine-tune-seg-comparison-graph}
\end{figure}

The IL-Score comparison between RBACA, CASA, and SeqFineTune on the selected combinations is illustrated in Fig. \ref{fine-tune-seg-comparison-graph}. The detailed results are reported in Table \ref{fine-tune-seg-comparison} and in the Appendix, in Table \ref{fine-tune-seg-comparison2}. SeqFineTune is used as a comparison method; its configuration remains constant, including the absence of memory and infinite $\beta$.


\begin{table*}[t]
\centering
\caption{Segmentation results for RBACA, CASA, and SeqFineTune across memory sizes and $\beta$.}
\label{fine-tune-seg-comparison}
\resizebox{\textwidth}{!}{%
\begin{tabular}{|cccc|cccc|}
\hline
\multicolumn{4}{|c|}{Setup}               & \multicolumn{4}{c|}{Results}                                                                       \\ \hline
\rowcolor[HTML]{B7B7B7} 
\begin{tabular}[c]{@{}c@{}}Fig. \ref{fine-tune-seg-comparison-graph}\\ Ref.\end{tabular} & MM     & Pruning & AL Method  & BWT                     & FWT                    & DSC                    & IL-Score               \\ \hline
S         & -      & -       & -          & $-$0.032 ± 0.008          & 0.493 ± 0.020          & 0.554 ± 0.009          & 0.338 ± 0.010          \\ \hline
\rowcolor[HTML]{EFEFEF} 
R11       & DM-3   & K-Means & Perf       & $-$0.005 ± 0.004          & \textbf{0.583 ± 0.002} & \textbf{0.727 ± 0.002} & \textbf{0.435 ± 0.003} \\
C11       & Static & LRU     & Perf       & \textbf{0.002 ± 0.001}  & 0.557 ± 0.002          & 0.709 ± 0.002          & 0.423 ± 0.002          \\ \hline
\rowcolor[HTML]{EFEFEF} 
R41       & Static & DBSCAN  & Perf       & 0.001 ± 0.005           & \textbf{0.581 ± 0.001} & \textbf{0.741 ± 0.003} & \textbf{0.441 ± 0.003} \\
C41       & Static & LRU     & Perf       & \textbf{0.003 ± 0.004}  & 0.577 ± 0.001          & 0.729 ± 0.005          & 0.436 ± 0.004          \\ \hline
\rowcolor[HTML]{EFEFEF} 
R81       & Static & LRU     & Perf       & \textbf{0.012 ± 0.003}  & \textbf{0.579 ± 0.000} & \textbf{0.743 ± 0.002} & \textbf{0.445 ± 0.002} \\
C81       & Static & LRU     & Perf       & \textbf{0.012 ± 0.003}  & \textbf{0.579 ± 0.000} & \textbf{0.743 ± 0.002} & \textbf{0.445 ± 0.002} \\ \hline
\rowcolor[HTML]{EFEFEF} 
R12       & DM-3   & KU      & $U_{th}$=0.0225 & $-$0.002 ± 0.001          & \textbf{0.577 ± 0.001} & \textbf{0.719 ± 0.003} & \textbf{0.431 ± 0.002} \\
C12       & Static & LRU     & Perf       & \textbf{$-$0.001 ± 0.002} & 0.566 ± 0.002          & 0.700 ± 0.005          & 0.422 ± 0.003          \\ \hline
\rowcolor[HTML]{EFEFEF} 
R42       & DM-3   & KU      & Perf       & \textbf{0.008 ± 0.007}  & 0.578 ± 0.000          & \textbf{0.739 ± 0.005} & \textbf{0.442 ± 0.005} \\
C42       & Static & LRU     & Perf       & $-$0.007 ± 0.001          & \textbf{0.581 ± 0.001} & 0.734 ± 0.001          & 0.436 ± 0.001          \\ \hline
\rowcolor[HTML]{EFEFEF} 
R82       & DM-3   & EGL     & Perf       & \textbf{0.001 ± 0.001}  & \textbf{0.582 ± 0.000} & \textbf{0.736 ± 0.000} & \textbf{0.440 ± 0.001} \\
C82       & Static & LRU     & Perf       & $-$0.013 ± 0.009          & 0.562 ± 0.003          & 0.718 ± 0.006          & 0.422 ± 0.006          \\ \hline
\rowcolor[HTML]{EFEFEF} 
R13       & Static & DBSCAN  & $U_{th}$=0.02   & 0.004 ± 0.002           & \textbf{0.562 ± 0.001} & \textbf{0.711 ± 0.005} & \textbf{0.426 ± 0.003} \\
C13       & Static & LRU     & Perf       & \textbf{0.005 ± 0.002}  & 0.557 ± 0.003          & 0.690 ± 0.008          & 0.417 ± 0.005          \\ \hline
\rowcolor[HTML]{EFEFEF} 
R43       & DM-3   & KU      & Perf       & \textbf{0.008 ± 0.003}  & \textbf{0.582 ± 0.002} & \textbf{0.739 ± 0.002} & \textbf{0.443 ± 0.002} \\
C43       & Static & LRU     & Perf       & 0.006 ± 0.002           & 0.575 ± 0.002          & 0.731 ± 0.002          & 0.437 ± 0.002          \\ \hline
\rowcolor[HTML]{EFEFEF} 
R83       & DM-3   & LRU     & Perf       & \textbf{0.010 ± 0.002}  & \textbf{0.582 ± 0.001} & \textbf{0.742 ± 0.001} & \textbf{0.445 ± 0.001} \\
C83       & Static & LRU     & Perf       & 0.003 ± 0.002           & 0.575 ± 0.002          & 0.732 ± 0.001          & 0.437 ± 0.002          \\ \hline
\end{tabular}
}
\end{table*}

\begin{figure*}[t]
\begin{minipage}[b]{\linewidth}
  \centering
  \centerline{\includegraphics[width=1\linewidth]{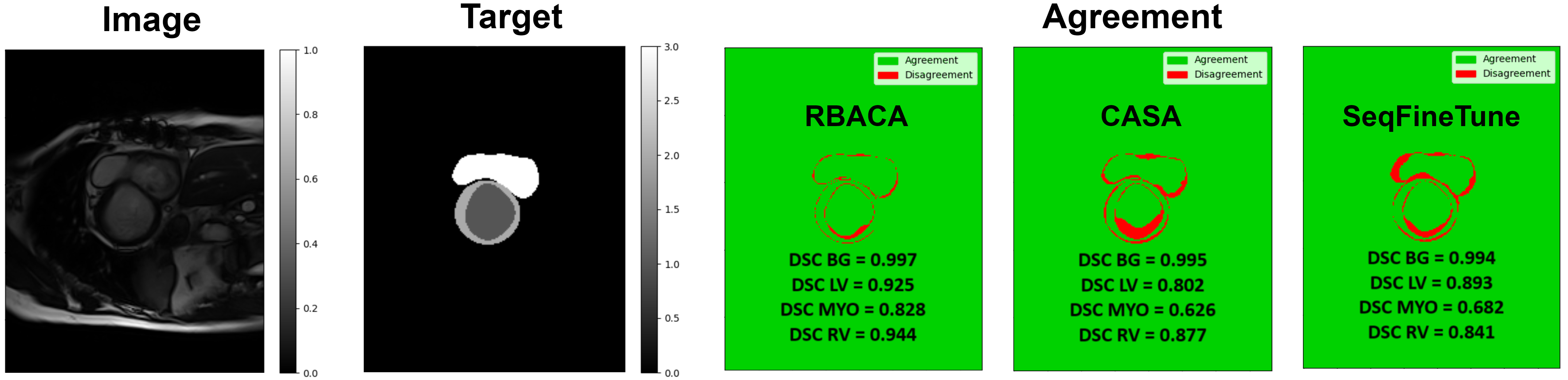}}
\end{minipage}
\caption{Image, target, and prediction agreement for a specific patient slice.}
\label{B9H8N8-pred-gt-agr}
\end{figure*}

SeqFineTune achieved the worst results. For each combination of $\beta$ and memory size, RBACA surpasses both CASA and SeqFineTune in performance.
The behavior of each model was observed for a specific patient. Fig. \ref{B9H8N8-pred-gt-agr} displays a cardiac slice, its target, and the agreement between each model's prediction and the target, along with the DSC for each class, demonstrating that RBACA achieved superior performance.

The memory distribution across PCs and pathologies was analyzed for both RBACA (R11) and CASA (C11), as illustrated in the Appendix, in Fig. \ref{seg_hist_qualitative}.
For both, 4 PCs were created: PC 0 contains data from Center 1 (Siemens), PC 1 from Center 5 (Canon), PC 2 from Centers 2 and 3 (both Philips), and PC 3 comprises a mix of all centers.
RBACA outperforms CASA, forming the same number of PCs with similar characteristics, and in this specific configuration, even with a smaller total memory $K_M$ ($160 < 200$ samples).





\subsection{Cardiac Classification}
\label{CardiacClassificationResultsandDiscussion}

The performance of ResNet-50 and ViT was assessed using RBACA, with and without DAug, under $\beta=430$, $K_M=128$, Static MM, Perf AL method and LRU pruning. The results are shown in the Appendix, in Table \ref{class_comparing_models_resnetwin}, indicating that ResNet-50 with DAug achieved the best performance.

ContextEval obtained FWT = 0.086 ± 0.016 and F1 Score = 0.086 ± 0.016 for this task. In each round, the overall F1 Score is the average of the F1 Scores for the classes present in the test labels. Additionally, F1 Score from a randomly initialized untrained model is 0 because it has no units in its final layer. The positive FWT indicates that training on a new episode benefits from having been previously trained on earlier episodes, even if those contain different classes.

\begin{figure}[t]
\begin{minipage}[b]{\linewidth}
  \centering
  \centerline{\includegraphics[width=1\linewidth]{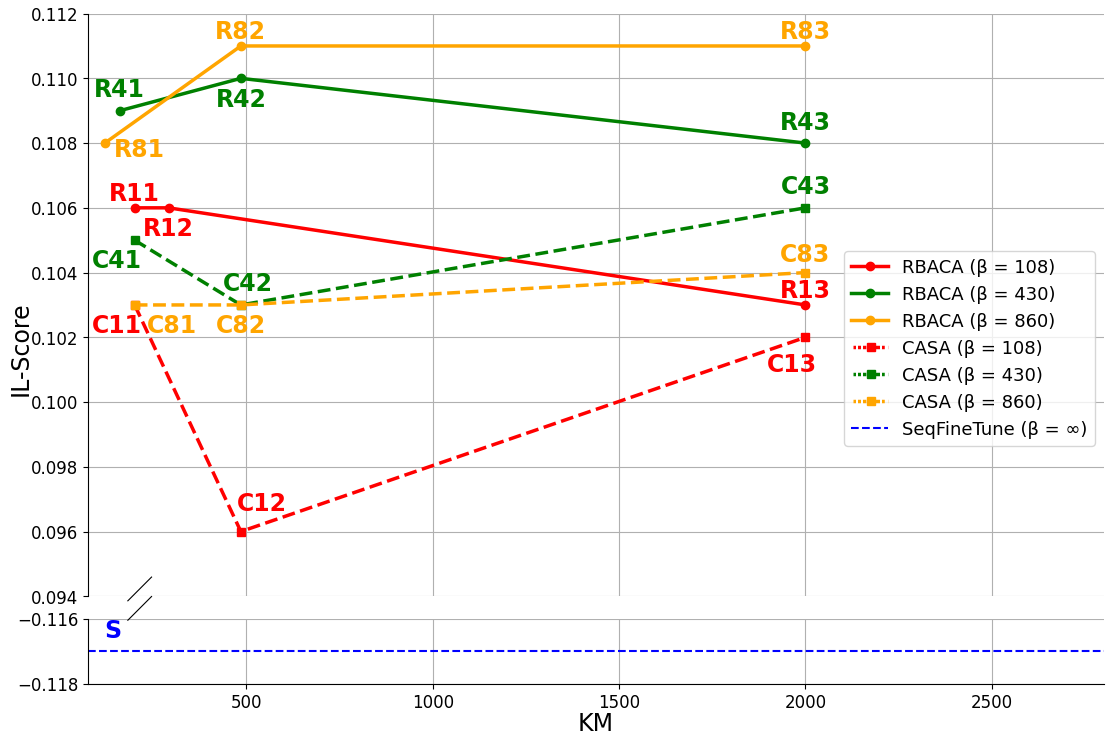}}
\end{minipage}
\caption{RBACA, CASA and SeqFineTune classification IL-Score for multiple configurations.}
\label{fine-tune-class-comparison-graph}
\end{figure}

SeqFineTune results are reported in Table \ref{fine-tune-class-comparison}.
The positive FWT demonstrates the benefit of using transfer learning. However, the negative BWT indicates catastrophic forgetting, as performance on a context drops significantly after learning from other contexts because the model tends to predict only the classes from the current training context.
For each of the 9 configurations used in segmentation, RBACA was tested with MM: \{Static, DM-3\}; Pruning: \{LRU, K-Means, GMMs, DBSCAN, Uncertainty, EGL, KU, EGLGMM\}; and AL: \{Perf, $U_{th} = \{0.025, 0.0225, 0.02\}$\}.
The performance comparison for RBACA, CASA and SeqFineTune across the combinations is shown in Fig. \ref{fine-tune-class-comparison-graph}. Detailed results are reported in Table \ref{fine-tune-class-comparison} and in the Appendix, in Table \ref{fine-tune-class-comparison2}. SeqFineTune has the poorest IL-Score performance, and RBACA outperforms CASA, which is consistent with the conclusions from segmentation, demonstrating RBACA's adaptability to different problems. 
However, in this problem, RBACA does not achieve high performance, and further advancements are needed to meet the standards required for medical applications.


\begin{table*}[t]
\centering
\caption{Classification results for RBACA, CASA, and SeqFineTune across memory sizes and $\beta$.}
\label{fine-tune-class-comparison}
\resizebox{\textwidth}{!}{%
\begin{tabular}{|cccc|cccc|}
\hline
\multicolumn{4}{|c|}{Setup}                                                                                                      & \multicolumn{4}{c|}{Results}                                                                      \\ \hline
\rowcolor[HTML]{B7B7B7} 
\begin{tabular}[c]{@{}c@{}}Fig. \ref{fine-tune-class-comparison-graph}\\ Ref.\end{tabular} & MM     & Pruning     & AL Method & BWT                    & FWT                    & F1 Score               & IL-Score               \\ \hline
S                                                   & -      & -           & -                                                   & $-$0.733 ± 0.028         & 0.074 ± 0.013          & 0.308 ± 0.031          & $-$0.117 ± 0.018         \\ \hline
\rowcolor[HTML]{EFEFEF} 
R11                                                 & Static & EGL         & Perf                                                & 0.025 ± 0.002          & \textbf{0.178 ± 0.003} & \textbf{0.114 ± 0.002} & \textbf{0.106 ± 0.002} \\
C11                                                 & Static & LRU         & Perf                                                & \textbf{0.037 ± 0.007} & 0.165 ± 0.001          & 0.107 ± 0.003          & 0.103 ± 0.004          \\ \hline
\rowcolor[HTML]{EFEFEF} 
R41                                                 & DM-3   & EGLGMM    & $U_{th}$=0.02                                            & 0.027 ± 0.002          & \textbf{0.182 ± 0.003} & \textbf{0.118 ± 0.003} & \textbf{0.109 ± 0.003} \\
C41                                                 & Static & LRU         & Perf                                                & \textbf{0.041 ± 0.005} & 0.166 ± 0.003          & 0.109 ± 0.004          & 0.105 ± 0.004          \\ \hline
\rowcolor[HTML]{EFEFEF} 
R81                                                 & DM-3   & K-Means     & $U_{th}$=0.02                                            & 0.027 ± 0.005          & \textbf{0.179 ± 0.001} & \textbf{0.118 ± 0.001} & \textbf{0.108 ± 0.003} \\
C81                                                 & Static & LRU         & Perf                                                & \textbf{0.032 ± 0.008} & 0.170 ± 0.006          & 0.107 ± 0.006          & 0.103 ± 0.007          \\ \hline
\rowcolor[HTML]{EFEFEF} 
R12                                                 & DM-3   & LRU         & Perf                                                & 0.025 ± 0.009          & \textbf{0.178 ± 0.001} & \textbf{0.114 ± 0.004} & \textbf{0.106 ± 0.006} \\
C12                                                 & Static & LRU         & Perf                                                & \textbf{0.027 ± 0.009} & 0.163 ± 0.006          & 0.098 ± 0.007          & 0.096 ± 0.007          \\ \hline
\rowcolor[HTML]{EFEFEF} 
R42                                                 & DM-3   & EGL         & Perf                                                & 0.027 ± 0.005          & \textbf{0.184 ± 0.002} & \textbf{0.119 ± 0.003} & \textbf{0.110 ± 0.004} \\
C42                                                 & Static & LRU         & Perf                                                & \textbf{0.040 ± 0.010} & 0.164 ± 0.004          & 0.104 ± 0.006          & 0.103 ± 0.007          \\ \hline
\rowcolor[HTML]{EFEFEF} 
R82                                                 & Static & K-Means     & Perf                                                & \textbf{0.035 ± 0.007} & \textbf{0.179 ± 0.003} & \textbf{0.118 ± 0.001} & \textbf{0.111 ± 0.004} \\
C82                                                 & Static & LRU         & Perf                                                & 0.035 ± 0.014          & 0.168 ± 0.003          & 0.105 ± 0.004          & 0.103 ± 0.009          \\ \hline
\rowcolor[HTML]{EFEFEF} 
R13                                                 & Static & Uncertainty & $U_{th}$=0.0225                                          & 0.016 ± 0.005          & \textbf{0.181 ± 0.001} & \textbf{0.111 ± 0.003} & \textbf{0.103 ± 0.003} \\
C13                                                 & Static & LRU         & Perf                                                & \textbf{0.034 ± 0.014} & 0.166 ± 0.004          & 0.105 ± 0.008          & 0.102 ± 0.010          \\ \hline
\rowcolor[HTML]{EFEFEF} 
R43                                                 & Static & EGL         & $U_{th}$=0.02                                            & 0.024 ± 0.005          & \textbf{0.181 ± 0.002} & \textbf{0.118 ± 0.004} & \textbf{0.108 ± 0.004} \\
C43                                                 & Static & LRU         & Perf                                                & \textbf{0.042 ± 0.003} & 0.167 ± 0.004          & 0.109 ± 0.003          & 0.106 ± 0.003          \\ \hline
\rowcolor[HTML]{EFEFEF} 
R83                                                 & Static & EGL         & Perf                                                & 0.030 ± 0.003          & \textbf{0.183 ± 0.003} & \textbf{0.119 ± 0.001} & \textbf{0.111 ± 0.003} \\
C83                                                 & Static & LRU         & Perf                                                & \textbf{0.036 ± 0.003} & 0.169 ± 0.001          & 0.108 ± 0.002          & 0.104 ± 0.002          \\ \hline
\end{tabular}
}
\end{table*}

\begin{figure}[t]
\begin{minipage}[b]{\linewidth}
  \centering
  \centerline{\includegraphics[width=1\linewidth]{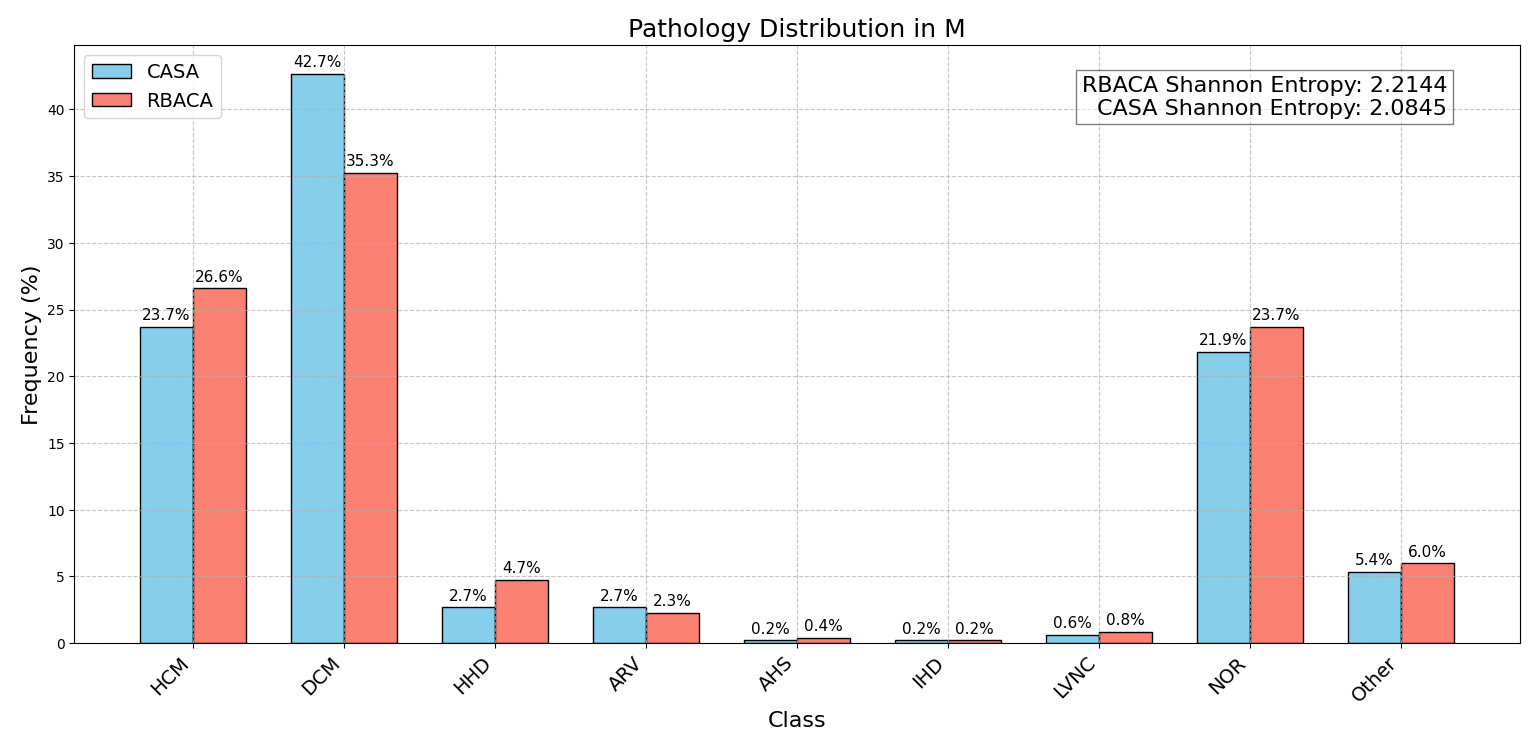}}
\end{minipage}
\caption{Pathology memory distribution in classification example.}
\label{class_hist_qualitative}
\end{figure}

The pathologies stored in memory for RBACA (R82) and CASA (C82) are presented in Fig. \ref{class_hist_qualitative}. RBACA achieves a better balance between classes and greater class representativeness in memory, as demonstrated by higher entropy \cite{shannon1948mathematical}, leading to superior performance than CASA.





%% file: sections/S6_conclusions.tex
\section{Conclusions and Future Work}

\label{conclusionss}

This work introduces a novel CAL framework for robust medical image analysis that enhances adaptability and generalizability while reducing annotation effort. 
A novel approach to evaluate CAL methods is established using a defined metric denominated IL-Score, which allows for the simultaneous assessment of transfer learning, forgetting, and final model performance.
The performance was evaluated in cardiac segmentation and pathology classification within Domain-IL and Class-IL scenarios, respectively. 
RBACA outperformed a baseline framework without CAL (SeqFineTune) and a state-of-the-art CAL method (CASA) across various memory sizes and annotation budgets in both scenarios, demonstrating its adaptability.
RBACA offers a range of adjustable features for each task, memory size, and annotation budget. It supports both static and dynamic memory strategies and employs more effective pruning. Additionally, it enhances representations using an AL method that considers informativeness.

The results underscore the significance of CAL in medical imaging. However, there is potential for further improvements. A promising research direction is to combine techniques from different CL families to mitigate their respective drawbacks and enhance their capabilities.
Another suggestion is to evaluate state-of-the-art methods across a range of problems within Domain-IL, Class-IL, and Task-IL scenarios. A potential approach for adapting rehearsal techniques to Task-IL
is to utilize multiple task-specific networks while maintaining a single rehearsal memory shared across tasks. 

CAL methods with storage mechanisms raise data privacy concerns, particularly when dealing with sensitive or proprietary data. An interesting research direction is to explore solutions that incorporate privacy-preserving representations. In the case of RBACA, a feasible approach could involve storing a latent representation of each PC instead of raw data. 

DL models require a large number of labeled samples for effective training, which can be problematic for CAL methods if the base training dataset is too small. A possible solution is to assign pseudo-labels to high-confidence samples, thereby expanding the base training set so that the CAL method can begin processing the continuous data stream after being effectively trained for the base context.  This enhances the model’s transfer learning capabilities to new contexts.

%% file: sections/attachments.tex
\section{}

\begin{table}[htb]
\centering
\caption{RBACA classification results with $\beta=430$, $K_M=128$, Static MM, Perf AL method and LRU pruning, for ResNet-50 and ViT, with and without data augmentation (DAug).}
\label{class_comparing_models_resnetwin}
\scalebox{0.88}{
\begin{tabular}{|c|c|c|c|c|}
\hline
\rowcolor[HTML]{999999} 
Model                                                    & BWT                    & FWT                    & F1 Score               & IL-Score                    \\ \hline
\begin{tabular}[c]{@{}c@{}}ResNet-50\\ DAug\end{tabular} & \textbf{0.041 ± 0.002} & \textbf{0.165 ± 0.002} & \textbf{0.107 ± 0.002} & \textbf{0.104 ± 0.002} \\ \hline
ResNet-50                                                & 0.033 ± 0.005          & 0.163 ± 0.002          & 0.105 ± 0.004          & 0.100 ± 0.004          \\ \hline
ViT                                                      & 0.005 ± 0.007          & 0.144 ± 0.001          & 0.097 ± 0.004          & 0.082 ± 0.005          \\ \hline
\begin{tabular}[c]{@{}c@{}}ViT\\ DAug\end{tabular}       & 0.003 ± 0.004          & 0.138 ± 0.003          & 0.096 ± 0.004          & 0.079 ± 0.004          \\ \hline
\end{tabular}%
}
\end{table}

\begin{table*}[htb]
\centering
\caption{Segmentation $\beta$ and memory configurations for RBACA and CASA.}
\label{fine-tune-seg-comparison2}
\scalebox{0.9}{
\begin{tabular}{|
>{\columncolor[HTML]{B7B7B7}}c |ccccccccc|ccccccccc|}
\hline
Framework                         & \multicolumn{9}{c|}{RBACA}                                                                                                                                                                                                     & \multicolumn{9}{c|}{CASA}                                                                                                                                                                                                      \\ \hline
Fig. \ref{fine-tune-seg-comparison-graph} Ref. & \multicolumn{1}{c|}{R11} & \multicolumn{1}{c|}{R41} & \multicolumn{1}{c|}{R81} & \multicolumn{1}{c|}{R12} & \multicolumn{1}{c|}{R42} & \multicolumn{1}{c|}{R82} & \multicolumn{1}{c|}{R13}  & \multicolumn{1}{c|}{R43}  & R83  & \multicolumn{1}{c|}{C11} & \multicolumn{1}{c|}{C41} & \multicolumn{1}{c|}{C81} & \multicolumn{1}{c|}{C12} & \multicolumn{1}{c|}{C42} & \multicolumn{1}{c|}{C82} & \multicolumn{1}{c|}{C13}  & \multicolumn{1}{c|}{C43}  & C83  \\ \hline
$\beta$                                 & \multicolumn{1}{c|}{108} & \multicolumn{1}{c|}{430} & \multicolumn{1}{c|}{860} & \multicolumn{1}{c|}{108} & \multicolumn{1}{c|}{430} & \multicolumn{1}{c|}{860} & \multicolumn{1}{c|}{108}  & \multicolumn{1}{c|}{430}  & 860  & \multicolumn{1}{c|}{108} & \multicolumn{1}{c|}{430} & \multicolumn{1}{c|}{860} & \multicolumn{1}{c|}{108} & \multicolumn{1}{c|}{430} & \multicolumn{1}{c|}{860} & \multicolumn{1}{c|}{108}  & \multicolumn{1}{c|}{430}  & 860  \\ \hline
$K_M$                                & \multicolumn{1}{c|}{160} & \multicolumn{1}{c|}{200} & \multicolumn{1}{c|}{200} & \multicolumn{1}{c|}{582} & \multicolumn{1}{c|}{388} & \multicolumn{1}{c|}{388} & \multicolumn{1}{c|}{2000} & \multicolumn{1}{c|}{2000} & 2400 & \multicolumn{1}{c|}{200} & \multicolumn{1}{c|}{200} & \multicolumn{1}{c|}{200} & \multicolumn{1}{c|}{485} & \multicolumn{1}{c|}{485} & \multicolumn{1}{c|}{485} & \multicolumn{1}{c|}{2000} & \multicolumn{1}{c|}{2000} & 2000 \\ \hline
$k$                                 & \multicolumn{1}{c|}{40}  & \multicolumn{1}{c|}{40}  & \multicolumn{1}{c|}{40}  & \multicolumn{1}{c|}{97}  & \multicolumn{1}{c|}{97}  & \multicolumn{1}{c|}{97}  & \multicolumn{1}{c|}{333}  & \multicolumn{1}{c|}{400}  & 400  & \multicolumn{1}{c|}{50}  & \multicolumn{1}{c|}{33}  & \multicolumn{1}{c|}{40}  & \multicolumn{1}{c|}{121} & \multicolumn{1}{c|}{97}  & \multicolumn{1}{c|}{80}  & \multicolumn{1}{c|}{500}  & \multicolumn{1}{c|}{333}  & 285  \\ \hline
no. PCs                           & \multicolumn{1}{c|}{4}   & \multicolumn{1}{c|}{5}   & \multicolumn{1}{c|}{5}   & \multicolumn{1}{c|}{6}   & \multicolumn{1}{c|}{4}   & \multicolumn{1}{c|}{4}   & \multicolumn{1}{c|}{6}    & \multicolumn{1}{c|}{5}    & 6    & \multicolumn{1}{c|}{4}   & \multicolumn{1}{c|}{6}   & \multicolumn{1}{c|}{5}   & \multicolumn{1}{c|}{4}   & \multicolumn{1}{c|}{5}   & \multicolumn{1}{c|}{6}   & \multicolumn{1}{c|}{4}    & \multicolumn{1}{c|}{6}    & 7    \\ \hline
\end{tabular}
}
\end{table*}

\begin{table*}[htb]
\centering
\caption{Classification $\beta$ and memory configurations for RBACA and CASA.}
\label{fine-tune-class-comparison2}
\scalebox{0.9}{
\begin{tabular}{|
>{\columncolor[HTML]{B7B7B7}}c |ccccccccc|ccccccccc|}
\hline
Framework                         & \multicolumn{9}{c|}{RBACA}                                                                                                                                                                                                     & \multicolumn{9}{c|}{CASA}                                                                                                                                                                                                      \\ \hline
Fig. \ref{fine-tune-class-comparison-graph} Ref. & \multicolumn{1}{c|}{R11} & \multicolumn{1}{c|}{R41} & \multicolumn{1}{c|}{R81} & \multicolumn{1}{c|}{R12} & \multicolumn{1}{c|}{R42} & \multicolumn{1}{c|}{R82} & \multicolumn{1}{c|}{R13}  & \multicolumn{1}{c|}{R43}  & R83  & \multicolumn{1}{c|}{C11} & \multicolumn{1}{c|}{C41} & \multicolumn{1}{c|}{C81} & \multicolumn{1}{c|}{C12} & \multicolumn{1}{c|}{C42} & \multicolumn{1}{c|}{C82} & \multicolumn{1}{c|}{C13}  & \multicolumn{1}{c|}{C43}  & C83  \\ \hline
$\beta$                                 & \multicolumn{1}{c|}{108} & \multicolumn{1}{c|}{430} & \multicolumn{1}{c|}{860} & \multicolumn{1}{c|}{108} & \multicolumn{1}{c|}{430} & \multicolumn{1}{c|}{860} & \multicolumn{1}{c|}{108}  & \multicolumn{1}{c|}{430}  & 860  & \multicolumn{1}{c|}{108} & \multicolumn{1}{c|}{430} & \multicolumn{1}{c|}{860} & \multicolumn{1}{c|}{108} & \multicolumn{1}{c|}{430} & \multicolumn{1}{c|}{860} & \multicolumn{1}{c|}{108}  & \multicolumn{1}{c|}{430}  & 860  \\ \hline
$K_M$                                & \multicolumn{1}{c|}{200} & \multicolumn{1}{c|}{160} & \multicolumn{1}{c|}{120} & \multicolumn{1}{c|}{291} & \multicolumn{1}{c|}{485} & \multicolumn{1}{c|}{485} & \multicolumn{1}{c|}{2000} & \multicolumn{1}{c|}{2000} & 2000 & \multicolumn{1}{c|}{200} & \multicolumn{1}{c|}{200} & \multicolumn{1}{c|}{200} & \multicolumn{1}{c|}{485} & \multicolumn{1}{c|}{485} & \multicolumn{1}{c|}{485} & \multicolumn{1}{c|}{2000} & \multicolumn{1}{c|}{2000} & 2000 \\ \hline
$k$                                 & \multicolumn{1}{c|}{50}  & \multicolumn{1}{c|}{40}  & \multicolumn{1}{c|}{40}  & \multicolumn{1}{c|}{97}  & \multicolumn{1}{c|}{97}  & \multicolumn{1}{c|}{121} & \multicolumn{1}{c|}{666}  & \multicolumn{1}{c|}{400}  & 400  & \multicolumn{1}{c|}{50}  & \multicolumn{1}{c|}{40}  & \multicolumn{1}{c|}{40}  & \multicolumn{1}{c|}{161} & \multicolumn{1}{c|}{121} & \multicolumn{1}{c|}{97}  & \multicolumn{1}{c|}{666}  & \multicolumn{1}{c|}{400}  & 333  \\ \hline
no. PCs                           & \multicolumn{1}{c|}{4}   & \multicolumn{1}{c|}{4}   & \multicolumn{1}{c|}{3}   & \multicolumn{1}{c|}{3}   & \multicolumn{1}{c|}{5}   & \multicolumn{1}{c|}{4}   & \multicolumn{1}{c|}{3}    & \multicolumn{1}{c|}{5}    & 5    & \multicolumn{1}{c|}{4}   & \multicolumn{1}{c|}{5}   & \multicolumn{1}{c|}{5}   & \multicolumn{1}{c|}{3}   & \multicolumn{1}{c|}{4}   & \multicolumn{1}{c|}{5}   & \multicolumn{1}{c|}{3}    & \multicolumn{1}{c|}{5}    & 6    \\ \hline
\end{tabular}
}
\end{table*}

\begin{figure*}[h!]
\begin{minipage}[b]{\linewidth}
  \centering
  \centerline{\includegraphics[width=0.78\linewidth]{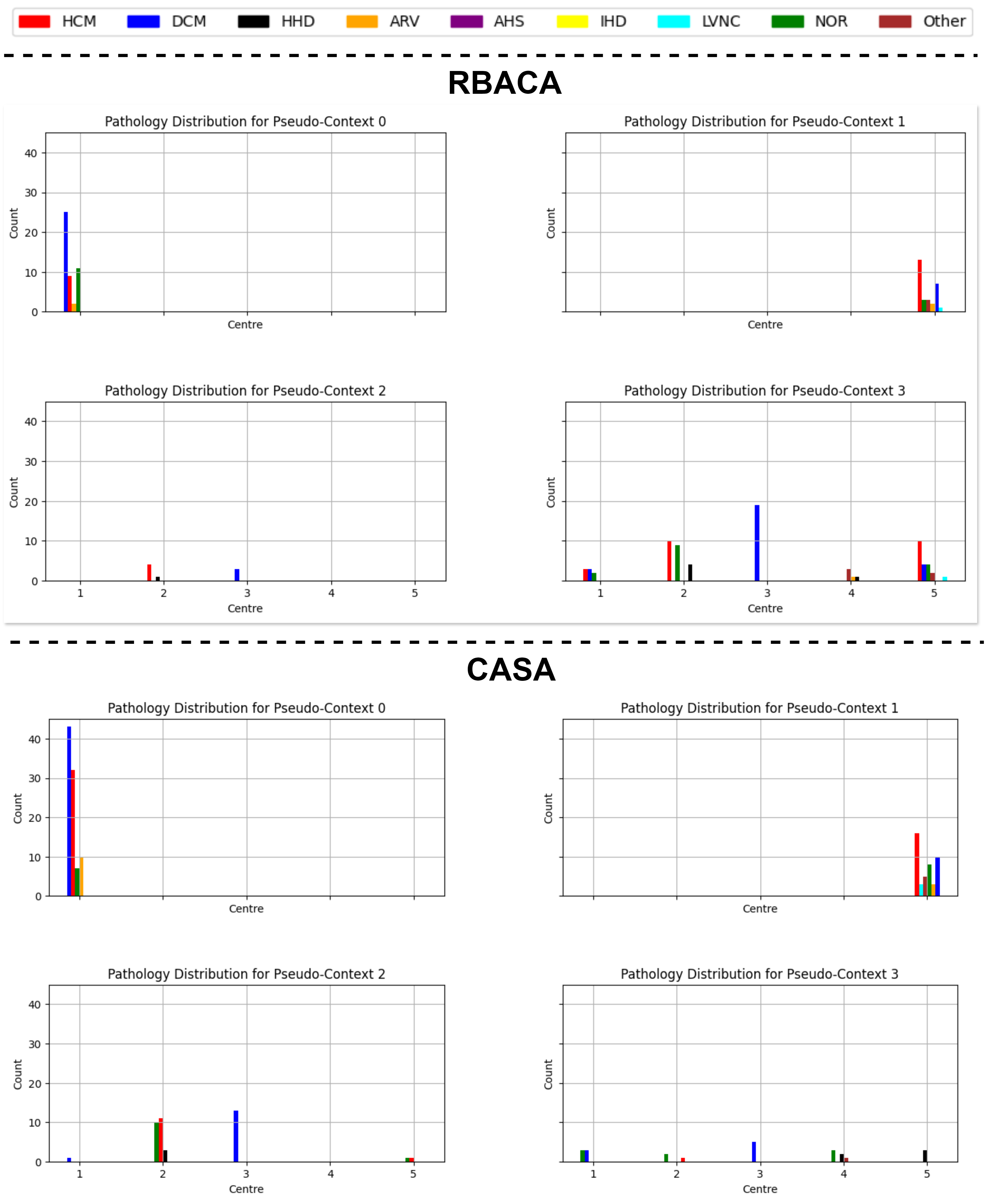}}
\end{minipage}
\caption{Pathology and center distribution per PC in segmentation example.}
\label{seg_hist_qualitative}
\end{figure*}